\documentclass{article}

     \usepackage[final]{neurips_2019}

\usepackage[utf8]{inputenc} 
\usepackage[T1]{fontenc}    
\usepackage{hyperref}       
\usepackage{url}            
\usepackage{booktabs}       
\usepackage{amsfonts}       
\usepackage{nicefrac}       
\usepackage{microtype}      
\usepackage{graphicx}
\usepackage{natbib}
\usepackage{amsmath}
\usepackage{amssymb}
\usepackage{mathtools}
\usepackage{wrapfig}
\usepackage{makecell}
\DeclareMathOperator*{\argmax}{argmax}
\DeclareMathOperator*{\argmin}{argmin}

\hypersetup{
  colorlinks  = true, 
  citecolor   = [rgb]{0,0.08,0.45},  
  linkcolor   = [rgb]{0,0.08,0.45}   
}

\title{Expert-guided Regularization \\ via Distance Metric Learning}

\author{%
  Shouvik Mani \qquad Mehdi Maasoumy \qquad Sina Pakazad \qquad Henrik Ohlsson\\ \\
  C3.ai \\
  Redwood City, CA \\
  \texttt{firstname.lastname@c3.ai} \\
}

\begin{document}

\maketitle

\begin{abstract}
High-dimensional prediction is a challenging problem setting for traditional statistical models. Although regularization improves model performance in high dimensions, it does not sufficiently leverage knowledge on feature importances held by domain experts. As an alternative to standard regularization techniques, we propose \textit{Distance Metric Learning Regularization (DMLreg)}, an approach for eliciting prior knowledge from domain experts and integrating that knowledge into a regularized linear model. First, we learn a Mahalanobis distance metric between observations from pairwise similarity comparisons provided by an expert. Then, we use the learned distance metric to place prior distributions on coefficients in a linear model. Through experimental results on a simulated high-dimensional prediction problem, we show that DMLreg leads to improvements in model performance when the domain expert is knowledgeable.
\end{abstract}

\section{Introduction}

Machine learning models exhibit poor predictive performance in high-dimensional problems, where the number of features is relatively large compared to the number of observations. In the high-dimensional setting, models suffer from high variance and overfit to the training data, a phenomenon known as the curse of dimensionality. Most methods used to mitigate this issue rely solely on data and fail to leverage knowledge from domain experts. In this paper, we propose a method that elicits an expert's knowledge on feature importances and integrates it into a regularized linear model to improve performance in high-dimensional settings.

High-dimensional datasets are ubiquitous in practice. Certain types of data such as images, text, and time series are inherently high-dimensional. In many domains, features are often cheap and numerous -- for instance, one may collect measurements from many sensors without knowing which ones are relevant to the prediction task at hand. In contrast, observations are often expensive to obtain and label. As a result, many problems have a large number of features, but relatively few observations. Under such circumstances, classical approaches are no longer valid \citep{Donoho}.

Given the wide presence of high-dimensional data, techniques such as regularization and dimensionality reduction are often used to fit simpler, less flexible models. These methods reduce the variance of the model at the cost of increased bias. For example, ridge regression and lasso are regularized variants of least squares linear regression, and they reduce the model’s variance by shrinking coefficient estimates toward zero \citep{Hoerl, lasso}.

\begin{figure*}[h]
\centering
\includegraphics[width=14cm]{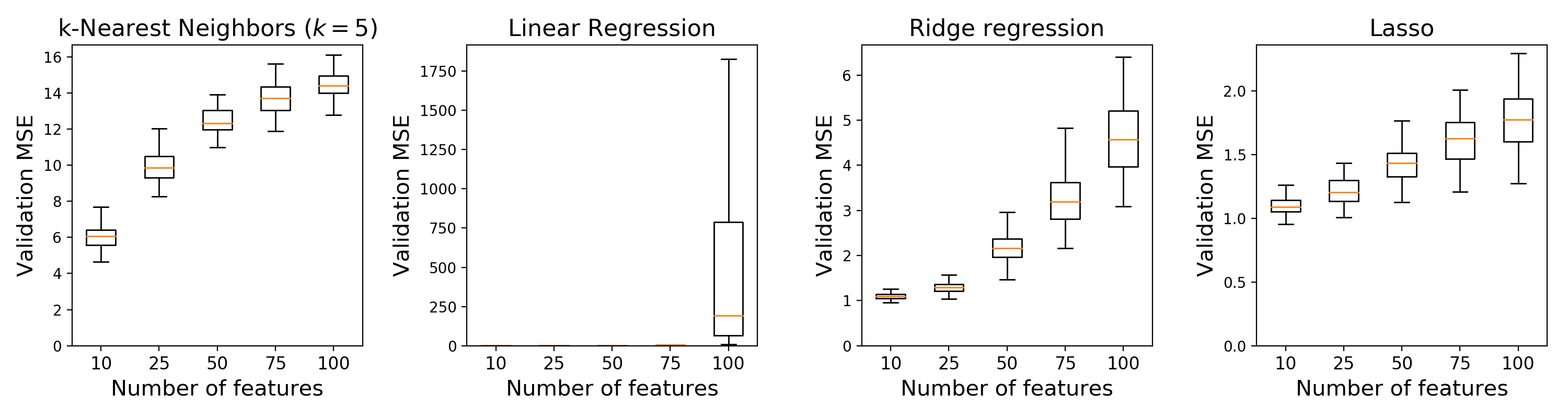}
\caption{Regularization is effective for high-dimensional problems: ridge regression and lasso (with optimal $\lambda$ chosen by cross-validation) outperform k-nearest neighbors and linear regression. Still, even ridge regression and lasso suffer from overfitting as the number of noise features increases.}
\vspace{-10pt}
\label{regularization}
\end{figure*}

Before introducing our approach, it is worth discussing the strengths and weaknesses of regularization. In Figure~\ref{regularization}, we compare the performance of k-nearest neighbors, linear regression, ridge regression, and lasso on a simulated regression problem where there are $n \in \{10, 25, 50, 75, 100\}$ features, out of which only the first $10$ features are truly associated with the response. The remaining $n - 10$ features are noise features with true coefficient values equal to zero. The training set consists of $m = 100$ observations and the validation MSE is evaluated on a validation set of $900$ observations\footnote{For complete details on the data generation procedure, see Section 4.}. Cross-validation is used for the ridge regression and lasso models to select the optimal regularization parameter from $\lambda \in \{10^{-4}, 10^{-3}, ..., 10^2\}$ at each $n$. As $n$ approaches $m$, the performance of k-nearest neighbors and linear regression rapidly deteriorate, but that of ridge regression and lasso remain relatively stable.

Still, even ridge regression and lasso suffer from overfitting, indicated by a steady increase in their validation MSEs as $n$ approaches $m$. With enough noise features, even regularization does not prevent overfitting, since chance associations between the features and the response on the training set result in some noise features being assigned nonzero coefficient estimates, even though those features are not truly associated with the response \citep{ISL}. Moreover, truly associated features with high true coefficient values have their coefficient estimates driven toward zero due to the regularization penalty.

To better understand these limitations and motivate our approach, it is helpful to look at regularization from a Bayesian perspective. Ridge regression and lasso have simple Bayesian interpretations \citep{MLPP, regularization-probabilistic}:
\begin{itemize}
  \item Ridge regression is the maximum a posteriori (MAP) solution from assuming a standard linear model with a Gaussian prior on the coefficients, with mean zero and variance parameterized by $\lambda$.
  \item Lasso is the MAP solution from assuming a standard linear model with a Laplace prior on the coefficients, with mean zero and scale parameterized by $\lambda$.
\end{itemize}

In both cases, all model coefficients share the same prior distribution, since they are all centered at zero and parameterized by the same regularization parameter $\lambda$. If we expect coefficients to be small or sparse in general, then ridge regression and lasso provide effective ways to encode that prior knowledge into the model. But if we have more informative prior knowledge about individual feature importances, then ridge regression and lasso do not help to encode that knowledge.

We propose a new regularization technique called Distance Metric Learning Regularization (DMLreg) to elicit prior knowledge on feature importances and incorporate that knowledge into a linear (classification or regression) model. In situations where training data are limited yet domain experts possess a wealth of prior knowledge about the problem, DMLreg combines human-driven priors with data-driven parameter estimation to fit a better regularized model.

\section{Related work}

This paper relies on distance metric learning, a concept pioneered by Xing et al. \citeyearpar{Xing} to automatically learn a distance metric from data and user guidance. Distance metric learning algorithms optimize a distance function, most commonly the Mahalanobis distance, to respect the user’s notion of similarity between instances. As input to the algorithm, the user provides weak supervision through pairs of similar and dissimilar instances (e.g. “$x_i$ is similar/dissimilar to $x_j$”). Alternatively, the user can provide relative comparisons between instances (e.g. “$x_i$ is more similar to $x_j$ than to $x_k$”) \citep{distance-metric-relative-comparisons}. Surveys by Kulis \citeyearpar{kulis} and Bellet et al. \citeyearpar{bellet}  review the vast literature on distance metric learning. 

Typically, the learned distance metric is used as an alternative to Euclidean distance to improve the performance of nearest-neighbor or kernel regression methods \citep{weinberger-lmnn, weinberger-kernel-regression}; however, in this paper, we integrate the learned metric into a regularized linear model. This offers advantages over nearest-neighbor methods such as greater interpretability, robustness to an incorrect learned distance metric, and a better model fit when a linear model is more appropriate for the underlying data.

Our work has parallels to research in knowledge elicitation. Daee et al. \citeyearpar{Daee2017} have a similar objective to elicit and incorporate expert knowledge, but they do so by using expected information gain to identify the most important features on which to query expert feedback. Unlike their approach which elicits knowledge at the feature level, our approach elicits knowledge at the sample level. This gives us advantages in high-dimensional settings where individual features may not be interpretable (e.g. pixel intensities in an image), but comparing samples for similarity may be easier.

More generally, a thorough discussion on the challenges and approaches in high-dimensional problems is provided by Hastie et al. \citeyearpar{ESL} and James et al. \citeyearpar{ISL}. These resources describe the curse of dimensionality in both kernel methods and linear models. Additionally, Murphy \citeyearpar{MLPP} and Wasserman \citeyearpar{all-of-stats} offer good references for Bayesian inference, a key framework used in this paper.

\section{Methodology}

Our DMLreg regularization approach has two key steps: (i) elicit domain knowledge on feature importances and (ii) incorporate that knowledge into a linear model. We describe this process in detail and derive MAP estimates for two models (linear regression and logistic regression) combined with DMLreg.

\subsection{Eliciting domain knowledge}

The first objective of DMLreg is to elicit knowledge on feature importances held by the domain expert. Here, one may ask: why not simply \textit{ask} the expert which features are relevant to the task at hand? While ideal, this is impractical in the high-dimensional setting -- it may be overwhelming or even impossible for an expert to rank or select good features in a feature space with hundreds or thousands of features.

In contrast, in many problems, it is relatively easy for an expert to look at pairs of observations and determine whether they are similar or dissimilar. In DMLreg, we exploit this ability of experts to do pairwise similarity comparisons in order to learn their tacit knowledge about feature importances. In particular, we aim to learn a weighting of the features that reflect their relative importances, and we achieve this task through distance metric learning.

We use the original distance metric learning formulation \citep{Xing}, through which we learn a Mahalanobis distance metric $d_{A}(x_i, x_j)$ between points $x_i$ and $x_j$ that respects the expert’s notion of similarity between observations. In other words, the learned distance metric assigns small distances between observations that the expert considers similar and large distances between observations that the expert considers dissimilar.
\begin{align}
d_{A}(x_i, x_j) = \sqrt{(x_i - x_j)^T A (x_i-x_j)}
\end{align}

To learn this distance metric, the domain expert must provide weak supervision on pairwise similarities in the form of sets $S$ and $D$, each containing pairs of observations which the expert considers similar and dissimilar, respectively.
\begin{align}
S &= \{(x_i, x_j): x_i \text{ and } x_j \text{ should be similar}\} \\
D &= \{(x_i, x_j): x_i \text{ and } x_j \text{ should be dissimilar}\}
\end{align}

Learning the distance metric can then be posed as a convex optimization problem: find an optimal matrix $A$ to minimize the sum of squared distances between the pairs of points in set $S$ that the expert has identified as similar.
\begin{align}
\underset{A}{\text{minimize}} \quad
& \textstyle\sum_{(x_i, x_j) \in S} \|x_i - x_j\|^{2}_{A} \\
\text{subject to} \quad
& \textstyle\sum_{(x_i, x_j) \in D} \|x_i - x_j\|_{A} \geq 1, \\
& A \succeq 0, \\
& A_{ij} = 0, \quad i \neq j.
\end{align}

The constraint in (5) maintains a lower bound on the total of distances between the dissimilar points in $D$ and prevents the trivial solution to (4) of $A = 0$. Constraint (6) requires $A$ to be a positive semi-definite matrix so that it is a valid distance metric that satisfies non-negativity and triangle inequality. Finally, constraint (7) restricts $A$ to be a diagonal matrix, which will make it easier to integrate $A$ with the model.

The diagonal elements of the learned matrix $A$ represent a weighting of the features based on their relative importances as indicated by the expert (at least with regard to computing distances between observations). For simplicity, we will refer to $A$ as the “distance metric” in the following discussion.

\subsection{Incorporating domain knowledge into the model}

Once we have captured the expert’s knowledge on feature importances in the form of a learned distance metric $A$, DMLreg integrates this knowledge into a regularized linear model. A natural way to combine prior knowledge with data is through Bayesian inference. In Bayesian inference, we specify a prior probability distribution $P(\theta)$ and a likelihood function $P(y|\theta)$. Using Bayes’ theorem, we can write the posterior distribution $P(\theta|y)$ as proportional to the likelihood times the prior.
\begin{align}
P(\theta|y) = \frac{P(y|\theta) P(\theta)}{P(y)} \propto P(y|\theta) P(\theta)
\end{align}

Specifically, let us first consider the case of Bayesian linear regression, which assumes a linear model $y = \theta^T x + \epsilon$ with Gaussian errors $\epsilon$. This implies that $y$ is also a Gaussian centered at the regression line with a corresponding likelihood function $P(y|\theta, x)$.
\begin{align}
y &= \theta^T x + \epsilon, \quad \epsilon \sim \mathcal{N}(0, \sigma^2) \\
&\iff y|\theta,x \sim \mathcal{N}(\theta^T x, \sigma^2) \\
P(y|\theta,x) &= \frac{1}{\sqrt{2 \pi \sigma^2}} exp \Big(\frac{-(y - \theta^T x)^2}{2 \sigma^2} \Big)
\end{align}

The prior distribution $P(\theta)$ in Bayesian linear regression lets us express our knowledge about the coefficients $\theta$. DMLreg encodes the knowledge held in the learned distance metric $A$ through a Gaussian prior on each coefficient. Like ridge regression, the Gaussian priors are all centered at zero, since $A$ does not provide any information on the direction of the relationship between each feature and the response.

However, the diagonal elements of $A$ represent the relative importances of the features. For $i \in 1,…,n$, if $A_{ii}$ is large, then we can expect a priori that feature $i$ has a large positive or negative coefficient $\theta_i$; in other words, we can expect $\theta_i$ to have a large variance around zero. So, we encode this information through the variance of each coefficient’s prior. This leads to placing a Gaussian prior distribution on each coefficient $\theta_i$ with mean zero and variance $A_{ii}$, and a corresponding density function $P(\theta_{i})$.
\begin{align}
\theta_{i \in 1,...,n} &\sim \mathcal{N}(0, A_{ii}) \\
P(\theta_i) &= \frac{1}{\sqrt{2 \pi A_{ii}^2}} exp \Big(\frac{-\theta_i^2}{2 A_{ii}} \Big) 
\end{align}

With our likelihood function $P(y|\theta, x)$ and prior distributions $P(\theta_{i})$ defined, we can use Bayes’ theorem to calculate the posterior distribution of $\theta$. Since we are only interested in the coefficient point estimates and not the full distribution of $\theta$, we can use maximum a posteriori (MAP) estimation to find the posterior mode $\hat{\theta}_{MAP}$: the value of $\theta$ that is most likely given the data and the priors.
\begin{align}
\hat{\theta}_{MAP} &= \argmax_{\theta} P(\theta | y_1,...,y_m, x_1,...,x_m) \\
&= \argmax_{\theta} \frac{P(y_1,...,y_m | \theta, x_1,...,x_m) P(\theta)}{P(y_1,...,y_m)} \\
&= \argmax_{\theta} \prod_{i=1}^{m} P(y_i | \theta, x_i) \prod_{j=1}^{n} P(\theta_j) \\
&= \argmax_{\theta} \sum_{i=1}^{m} logP(y_i | \theta, x_i) + \sum_{j=1}^{n} logP(\theta_j) \\
&= \argmax_{\theta} \sum_{i=1}^{m} \frac{-(y_i - \theta^T x_i)^2}{2 \sigma^2} + \sum_{j=1}^{n} \frac{-\theta_j^2}{2 A_{jj}} \\
&= \argmax_{\theta} - \frac{1}{2 \sigma^2} \sum_{i=1}^{m} (y_i - \theta^T x_i)^2 - \frac{1}{2} \sum_{j=1}^{n} \frac{\theta_j^2}{A_{jj}} \\
&= \argmin_{\theta} \sum_{i=1}^{m} (y_i - \theta^T x_i)^2 + \sum_{j=1}^{n} \frac{\theta_j^2}{A_{jj}} \\
&= \argmin_{\theta} \|y - X \theta \|_{2}^{2} + \theta^T A^{-1} \theta
\end{align}

Line (21) is the loss function for linear regression with DMLreg (Gaussian prior). We can solve for $\hat{\theta}$ to obtain a closed-form solution for the coefficient estimates.
\begin{align}
\nabla_{\hat{\theta}} \Big(\|y - X \hat{\theta} \|_{2}^{2} + \hat{\theta}^T A^{-1} \hat{\theta} \Big) = 0\\
-2 X^T (y - X \hat{\theta}) + 2 A^{-1} \hat{\theta} = 0 \\
-X^T y + X^T X \hat{\theta} + A^{-1} \hat{\theta} = 0 \\
(X^T X + A^{-1}) \hat{\theta} = X^T y \\
\hat{\theta} = (X^T X + A^{-1})^{-1} X^T y
\end{align} 

Thus, for linear regression with DMLreg (Gaussian prior), we can compute coefficient estimates $\hat{\theta}$ in terms of $X, y$, and the learned distance metric $A$. This is equivalent to the coefficient estimates for ridge regression when $A= \lambda I$, for regularization parameter $\lambda$. An interpretation of this result is that linear regression with DMLreg (Gaussian prior) is a generalization of ridge regression with a separate regularization parameter $A_{ii}$ for each feature $i$, instead of a single regularization parameter $\lambda$ across all features.

Apart from linear regression, DMLreg can be easily applied to any generalized linear model. In Table~\ref{dmlreg-models-table}, we show how to fit linear regression and logistic regression models with DMLreg for both Gaussian and Laplace priors on the coefficients. \footnote{For the Logistic Regression + DMLreg models in Table 1, $\sigma(\cdot)$ is the Sigmoid function $\sigma(x) = \frac{1}{1 + exp(-x)}$.}

\begin{table}
  \caption{Fitting linear models with DMLreg using Gaussian and Laplace priors on coefficients. The diagonal matrix $A$ represents the learned distance metric from the first step.}
  \label{dmlreg-models-table}
  \centering
  \begin{tabular}{lccc}
    \toprule
    Model Name     &    Assumptions   &   Optimization Method \\
    \midrule
    \makecell[l]{Linear Regression \\+ DMLreg (Gaussian prior)}  &   \makecell{$\theta_{i} \sim \mathcal{N}(0, A_{ii})$ \\[0.1cm] $y \sim \mathcal{N}(\theta^T x, \sigma^2)$}   &  \makecell{Analytic solution: \\[0.1cm] $\theta = (X^T X + A^{-1})^{-1} X^T y$}  \\[0.6cm]
    \makecell[l]{Linear Regression\\+ DMLreg (Laplace prior)}   &   \makecell{$\theta_{i} \sim Laplace(0, A_{ii})$ \\[0.1cm] $y \sim \mathcal{N}(\theta^T x, \sigma^2)$}   &   \makecell{Subgradient method: \\[0.1cm] $\argmin_{\theta} \|y - X \theta \|_{2}^{2} + \|A^{-1} \theta\|_{1}$} \\[0.6cm]
    \makecell[l]{Logistic Regression\\+ DMLreg (Gaussian prior)}   &   \makecell{$\theta_{i} \sim \mathcal{N}(0, A_{ii})$ \\[0.1cm] $y \sim Bern(p=\sigma(\theta^T x))$}    &   \makecell{Gradient descent: \\[0.1cm] $\theta := \theta + \alpha [y - \sigma(X \theta)] X - \alpha A^{-1} \theta$} \\[0.6cm]
    \makecell[l]{Logistic Regression\\+ DMLreg (Laplace prior)}   &   \makecell{$\theta_{i} \sim Laplace(0, A_{ii})$ \\[0.1cm] $y \sim Bern(p=\sigma(\theta^T x))$}    &   \makecell{Subgradient method: \\[0.1cm] $\argmin_{\theta} -y log \sigma(X \theta) -$ \\ $(1 - y) log [1 - \sigma(X \theta)] + \|A^{-1} \theta\|_{1}$} \\
    \bottomrule
  \end{tabular}
\end{table}

\section{Results}

\begin{figure*}[t]
\centering
\includegraphics[width=14cm]{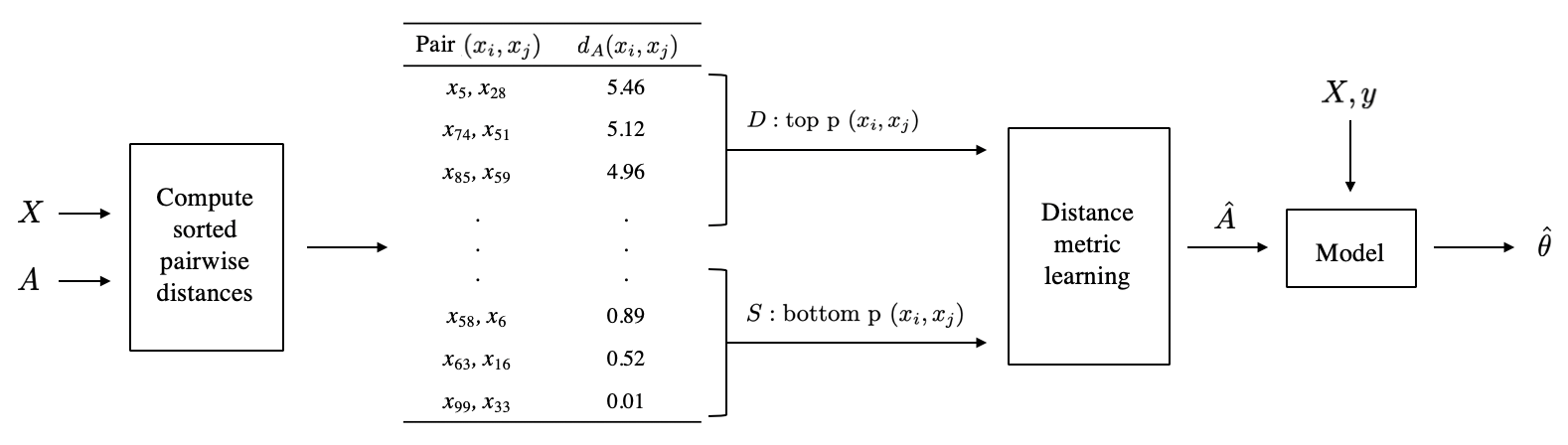}
\caption{Experiment used to evaluate DMLreg on artificially-generated data $X, y$ using simulated domain knowledge in $A$.}
\label{experiment-setup}
\end{figure*}

We evaluate DMLreg through an experiment on an artificial dataset with simulated domain knowledge. The experiment is outlined in Figure~\ref{experiment-setup}.

First, we generate artificial high-dimensional datasets with $m=100$ observations and $n \in \{10, 25, 50, 75, 100\}$ features. We sample an $n$-dimensional coefficient vector $\theta$ with the first 10 coefficients drawn from a Uniform distribution between -10 and 10 and the remaining $n-10$ coefficients set exactly to zero (line 27). Keeping $\theta$ fixed, we simulate 100 datasets $X, y$ through the procedure in lines 28-30. To be precise, our training set has $m = 100$ observations; we hold out a separate validation set with 900 observations.
\begin{align}
\theta &\in \mathbb{R}^n, \quad \theta_{1,...,10} \sim \textit{Unif}(-10, 10), \quad \theta_{11,...n} = 0 \\
X &\in \mathbb{R}^{m \times n}, \quad X \sim \textit{Unif}(0, 1)\\
\epsilon &\in \mathbb{R}^{m}, \quad \epsilon \sim \mathcal{N}(0, 1) \\
y &\in \mathbb{R}^{m}, \quad y = X \theta + \epsilon
\end{align}

To simulate an expert’s domain knowledge on feature importances, we create three true distance metrics $A$, listed in Table~\ref{true-metrics-table}, which represent varying levels of expertise: perfect, noisy, and incorrect. These true metrics $A$ are used for computing pairwise distances between observations in order to generate sets $S$ and $D$ for distance metric learning (to simulate an expert manually comparing observations and creating these sets).
\begin{table}[h]
\caption{True metrics used in the experiment to simulate domain knowledge on feature importances with varying levels of expertise.}
\label{true-metrics-table}
\vskip 0.15in
\begin{center}
\begin{small}
\begin{sc}
\begin{tabular}{lcccr}
\toprule
Knowledge type & True metric value \\
\midrule
Perfect   &   $A = \textit{diag}(|\theta_1|, ..., |\theta_n|)$ \\[0.3cm]
Noisy     &   \begin{tabular}{@{}c@{}} $A = \textit{diag}(|\theta_1 + \epsilon_1|, ..., |\theta_n + \epsilon_n|)$, \quad $\epsilon_i \sim \mathcal{N}(0, 0.5)$ \end{tabular} \\[0.4cm]
Incorrect &   \begin{tabular}{@{}c@{}} $A = \textit{diag}(|\alpha_1|, ..., |\alpha_n|)$, \quad $\alpha_i \sim \textit{Unif}(-5, 5)$ \end{tabular} \\
\bottomrule
\end{tabular}
\end{sc}
\end{small}
\end{center}
\vskip -0.1in
\end{table}

In practice, we expect real-world expert knowledge to be similar to a noisy or incorrect knowledge metric. The noisy knowledge metric is a diagonal matrix which weighs each feature by the magnitude of its true coefficient value $\theta_i$ plus some error $\epsilon_i$. It represents an expert who knows the correct feature importances up to some error. In contrast, the incorrect knowledge metric weighs the features with completely random weights $\alpha_i$.

Now, we explain how to generate sets $S$ and $D$ in the experiment. For a feature matrix $X$ and a true metric $A$, we compute pairwise distances between all observation pairs $(x_i, x_j)$ in $X$ using the Mahalanobis distance $ d_{A}(x_i, x_j) = \sqrt{(x_i - x_j)^T A (x_i - x_j)}$. The $p$ pairs with the highest distances are added to set $D$, and the $p$ pairs with the lowest distances are added to set $S$. 

The sets $S$ and $D$ are then passed on to the distance metric learning algorithm to learn an approximation $\hat{A}$ for $A$. The learned metric $\hat{A}$, along with $X$ and $y$ are used as inputs to DMLreg, which fits a regularized linear model.

\begin{figure*}[t]
\centering
\includegraphics[width=14cm]{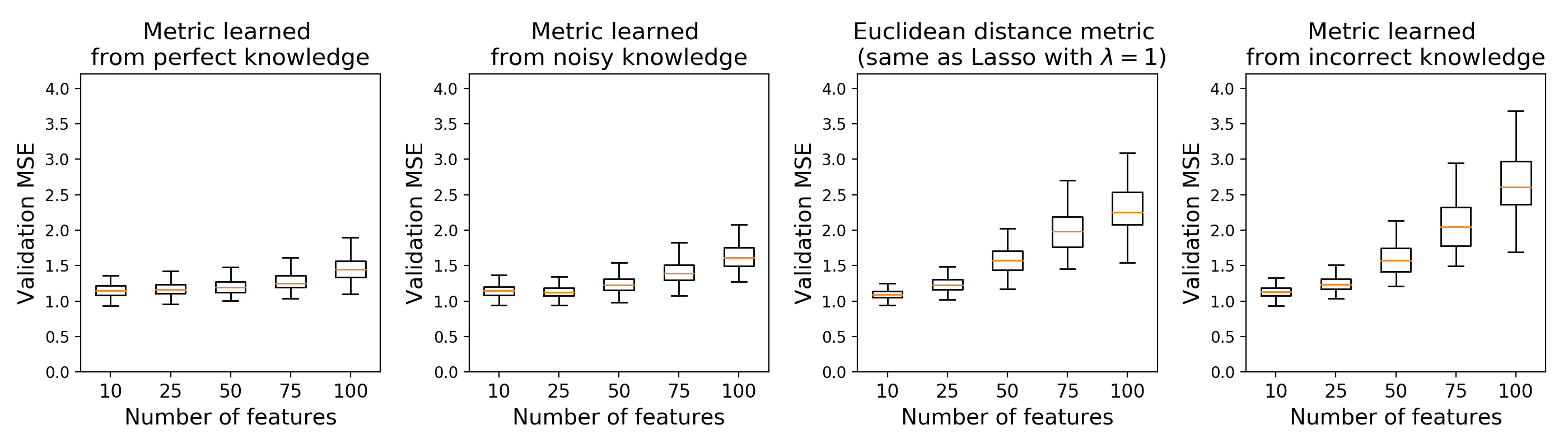}
\caption{Performance of Linear Regression + DMLreg (Laplace prior) model using various learned distance metrics as the number of features increases, for a fixed $m=100$ observations.}
\vspace{-10pt}
\label{dmlreg-1}
\end{figure*}

We evaluate the performance of the Linear Regression + DMLreg (Laplace prior) model on the validation set for each learned metric $\hat{A}$ and for each $n \in \{10, 25, 50, 75, 100\}$ features. Beside the metrics $\hat{A}$ learned to approximate the three true metrics $A$, we also consider the Euclidean distance metric $A = I$, which is identical to the lasso ($\lambda=1$) model when used in DMLreg with a Laplace prior. For this evaluation, we fix $p = 300$ pairs. The validation set MSEs are aggregated across the 100 $X, y$ datasets and displayed in Figure~\ref{dmlreg-1}.

For high values of $n$, DMLreg using metrics learned from perfect or noisy knowledge performs better than DMLreg using the Euclidean distance metric (lasso with $\lambda=1$). DMLreg substantially outperforms k-nearest neighbors and linear regression evaluated on the same data (refer to Figure~\ref{regularization}). It also performs slightly better than the lasso model which had $\lambda$ tuned through grid search and cross-validation (refer to Figure~\ref{regularization}), even though no hyperparameter search was required for its DMLreg counterpart. Nevertheless, when provided a metric learned from incorrect knowledge, DMLreg performs worse than lasso. These results suggest that DMLreg is effective for high-dimensional problems as long as the domain knowledge provided through the learned  metric is reasonably correct.

\begin{wrapfigure}[15]{r}{0.5\textwidth}
\centering
\vspace{-\intextsep}
\includegraphics[width=0.5\textwidth]{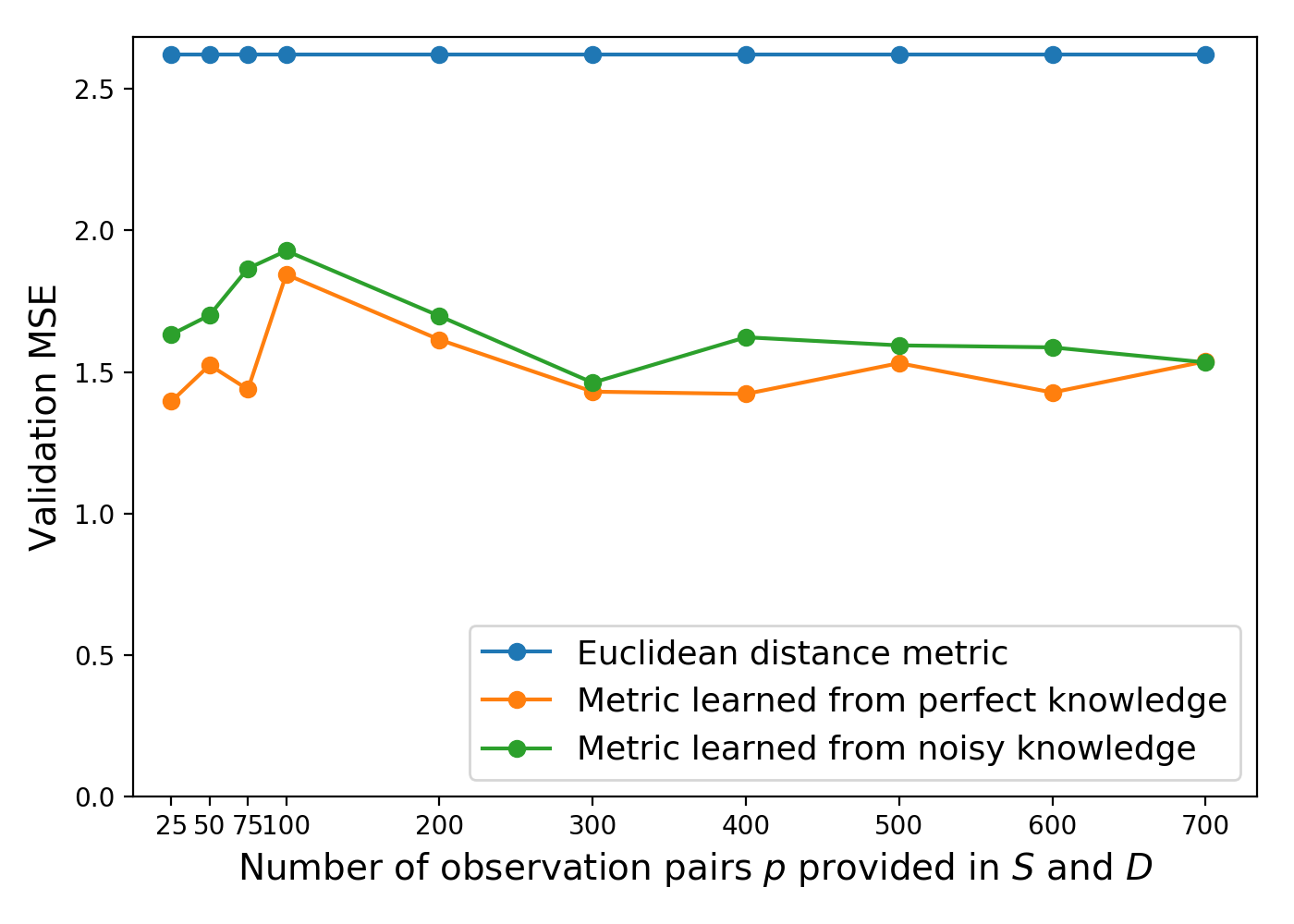}
\caption{DMLreg performance with $n=100$ features against the number of observation pairs $p$.}
\label{dmlreg-3}
\end{wrapfigure}
An important practical concern when using DMLreg is: how many pairs of observations $p$ does one need to provide in sets $S$ and $D$ in order to get good performance? Figure~\ref{dmlreg-3} shows the validation MSEs for DMLreg on a single dataset $X, y$ with a fixed $n=100$ features as $p$ is varied from 25 to 700 observations pairs. For DMLreg using the metric learned from noisy knowledge, providing just $p=25$ observation pairs yields nearly the same performance as $p=300$ pairs, as in the experiment above.

\begin{figure*}[t]
\centering
\includegraphics[width=14cm]{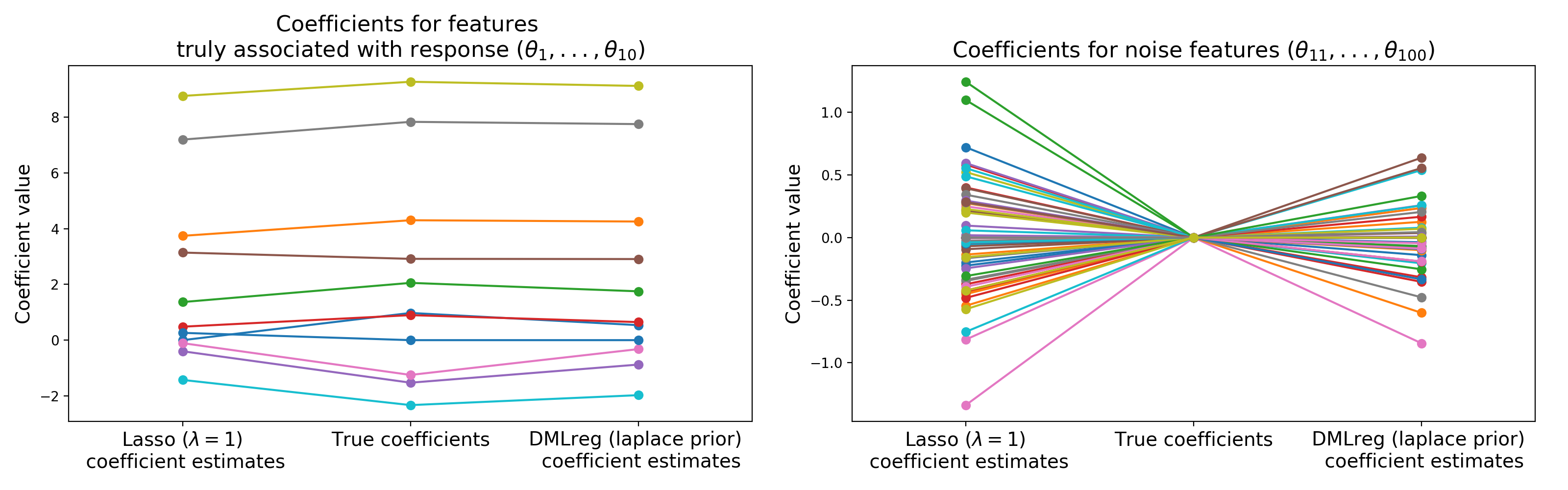}
\caption{True coefficients vs. coefficient estimates for the Linear Regression + DMLreg (Laplace prior) model and the Lasso ($\lambda=1$) model.}
\vspace{-10pt}
\label{dmlreg-coefficients}
\end{figure*}

To understand why the DMLreg model using the metric learned from noisy knowledge outperforms lasso, it is helpful to examine the coefficient estimates in Figure~\ref{dmlreg-coefficients}. For a single dataset $X, y$, Figure~\ref{dmlreg-coefficients} shows the coefficients for the first 10 (relevant) features on the left and the remaining (noise) features on the right. It is clear that the coefficient estimates for the DMLreg model (with noisy knowledge and Laplace prior) are closer to the true coefficients than the coefficient estimates for the lasso ($\lambda=1$) model. In fact, the lasso model slightly underestimates the coefficients of the truly associated features and overestimates the coefficients of the noise features when compared to the DMLreg model.

\section{Conclusion}
When training data is limited but domain knowledge on feature importances is available, DMLreg can be used to fit a better regularized model. Using distance metric learning and regularization, DMLreg elicits and integrates expert knowledge into a linear model. Through an experiment on artificial data using simulated domain knowledge, we demonstrated that DMLreg outperforms ridge regression and lasso when the knowledge elicited is approximately correct.

\bibliography{expert_guided_regularization_paper}

\begin{thebibliography}{15}
\providecommand{\natexlab}[1]{#1}
\providecommand{\url}[1]{\texttt{#1}}
\expandafter\ifx\csname urlstyle\endcsname\relax
  \providecommand{\doi}[1]{doi: #1}\else
  \providecommand{\doi}{doi: \begingroup \urlstyle{rm}\Url}\fi

\bibitem[Bellet et~al.(2013)Bellet, Habrard, and Sebban]{bellet}
Bellet, A., Habrard, A., and Sebban, M.
\newblock A survey on metric learning for feature vectors and structured data.
\newblock \emph{CoRR}, abs/1306.6709, 2013.

\bibitem[Daee et~al.(2017)Daee, Peltola, Soare, and Kaski]{Daee2017}
Daee, P., Peltola, T., Soare, M., and Kaski, S.
\newblock Knowledge elicitation via sequential probabilistic inference for
  high-dimensional prediction.
\newblock \emph{Machine Learning}, 106\penalty0 (9):\penalty0 1599--1620, Oct
  2017.

\bibitem[Donoho(2000)]{Donoho}
Donoho, D.~L.
\newblock High-dimensional data analysis: The curses and blessings of
  dimensionality.
\newblock In \emph{AMS Conference on Math Challenges of the 21st Century},
  2000.

\bibitem[Hastie et~al.(2009)Hastie, Tibshirani, and Friedman]{ESL}
Hastie, T., Tibshirani, R., and Friedman, J.
\newblock \emph{The Elements of Statistical Learning: Data Mining, Inference,
  and Prediction}.
\newblock Springer, 2009.

\bibitem[Hoerl \& Kennard(1970)Hoerl and Kennard]{Hoerl}
Hoerl, A.~E. and Kennard, R.~W.
\newblock Ridge regression: Biased estimation for nonorthogonal problems.
\newblock \emph{Technometrics}, 12\penalty0 (1):\penalty0 55--67, 1970.

\bibitem[James et~al.(2014)James, Witten, Hastie, and Tibshirani]{ISL}
James, G., Witten, D., Hastie, T., and Tibshirani, R.
\newblock \emph{An Introduction to Statistical Learning: With Applications in
  R}, pp.\  238--244.
\newblock Springer, 2014.

\bibitem[Keng(2016)]{regularization-probabilistic}
Keng, B.
\newblock A probabilistic interpretation of regularization, 2016.

\bibitem[Kulis(2013)]{kulis}
Kulis, B.
\newblock \emph{Metric Learning: A Survey}.
\newblock Foundations and trends in machine learning. Now Publishers, 2013.

\bibitem[Murphy(2012)]{MLPP}
Murphy, K.~P.
\newblock \emph{Machine Learning: A Probabilistic Perspective}.
\newblock The MIT Press, 2012.

\bibitem[Schultz \& Joachims(2003)Schultz and
  Joachims]{distance-metric-relative-comparisons}
Schultz, M. and Joachims, T.
\newblock Learning a distance metric from relative comparisons.
\newblock In \emph{Proceedings of the 16th International Conference on Neural
  Information Processing Systems}, 2003.

\bibitem[Tibshirani(1996)]{lasso}
Tibshirani, R.
\newblock Regression shrinkage and selection via the lasso.
\newblock \emph{Journal of the Royal Statistical Society. Series B
  (Methodological)}, 58\penalty0 (1):\penalty0 267--288, 1996.

\bibitem[Wasserman(2010)]{all-of-stats}
Wasserman, L.
\newblock \emph{All of Statistics: A Concise Course in Statistical Inference}.
\newblock Springer, 2010.

\bibitem[Weinberger \& Tesauro(2007)Weinberger and
  Tesauro]{weinberger-kernel-regression}
Weinberger, K.~Q. and Tesauro, G.
\newblock Metric learning for kernel regression.
\newblock In \emph{Artificial Intelligence and Statistics}, pp.\  612--619,
  2007.

\bibitem[Weinberger et~al.(2006)Weinberger, Blitzer, and Saul]{weinberger-lmnn}
Weinberger, K.~Q., Blitzer, J., and Saul, L.~K.
\newblock Distance metric learning for large margin nearest neighbor
  classification.
\newblock In \emph{Advances in neural information processing systems}, 2006.

\bibitem[Xing et~al.(2002)Xing, Ng, Jordan, and Russell]{Xing}
Xing, E.~P., Ng, A.~Y., Jordan, M.~I., and Russell, S.
\newblock Distance metric learning, with application to clustering with
  side-information.
\newblock In \emph{Proceedings of the 15th International Conference on Neural
  Information Processing Systems}, 2002.

\end{thebibliography}
\bibliographystyle{icml2019}

\end{document}